\documentclass[11pt]{article}
\usepackage{pifont}

\usepackage[final]{acl}

\usepackage{times}
\usepackage{latexsym}
\usepackage{subcaption}
\usepackage{amsmath}
\usepackage{amssymb}

\usepackage[T1]{fontenc}
\usepackage[utf8]{inputenc}

\usepackage{microtype}

\usepackage{inconsolata}

\usepackage{graphicx}

\usepackage{tcolorbox}
\usepackage{fvextra}

\fvset{
  breaksymbolleft={},
  breaksymbolright={}
}

\DefineVerbatimEnvironment{Prompt}{Verbatim}{
  breaklines,
  fontsize=\small,
  commandchars=\\\{\}
}

\tcbset{
  listing box/.style={
    colback=gray!10,
    colframe=black,
    boxrule=0.4pt,
    arc=0pt,
    left=4pt,
    right=4pt,
    top=4pt,
    bottom=4pt
  }
}

\title{Language Statistics and False Belief Reasoning: Evidence from 41 Open-Weight LMs}

\author{
  Sean Trott \\
  Rutgers University - Newark \\
  \texttt{sean.trott@rutgers.edu}
  \And
  Samuel Taylor \\
  UC San Diego \\
  \texttt{s6taylor@ucsd.edu}
  \And
  Cameron Jones \\
  Stony Brook University \\
  \texttt{cameron.jones@stonybrook.edu}
  \AND
  James A. Michaelov \\
  MIT \\
  \texttt{jamic@mit.edu}
  \And
  Pamela D. Rivi\`ere \\
  Rutgers University - Newark \\
  \texttt{pamela.riviereruiz@rutgers.edu}
}

\begin{document}
\maketitle
\begin{abstract}

Research on mental state reasoning in language models (LMs) has the potential to inform theories of human social cognition---such as the theory that mental state reasoning emerges in part from \textit{language exposure}---and our understanding of LMs themselves. Yet much published work on LMs relies on a relatively small \textit{sample} of closed-source LMs, limiting our ability to rigorously test psychological theories and evaluate LM capacities. Here, we replicate and extend published work on the \textit{false belief task} by assessing LM mental state reasoning behavior across 41 open-weight models (from distinct model families). We find \textit{sensitivity} to implied knowledge states in $34\%$ of the LMs tested; however, consistent with prior work, none fully ``explain away'' the effect in humans. Larger LMs show increased sensitivity and also exhibit higher psychometric predictive power. Finally, we use LM behavior to generate and test a novel hypothesis about human cognition: both humans and LMs show a bias towards attributing false beliefs when knowledge states are cued using a non-factive verb (``John thinks...'') than when cued indirectly (``John looks in the...''). Unlike the primary effect of knowledge states, where human sensitivity exceeds that of LMs, the magnitude of the human knowledge cue effect falls squarely within the distribution of LM effect sizes---suggesting that distributional statistics of language can in principle account for the latter but not the former in humans. These results demonstrate the value of using larger samples of open-weight LMs to test theories of human cognition and evaluate LM capacities. 

\end{abstract}

\section{Introduction}\label{sec:intro}

Advances in the performance of language models (LMs) have raised questions about which aspects of human cognition can emerge in a learner trained purely on the distributional statistics of language. Recent debates have focused on a range of topics, including grammatical knowledge \cite{dentella2023systematic, hu2024language}, analogical reasoning \citep{lewis2025evaluating, webb2023emergent}, and mental state reasoning \citep{nematzadeh_evaluating_2018, sap-etal-2022-neural, kosinski2024evaluating, trott2023large, hu2025re, shapira-etal-2024-clever, pang2025large, jones2024comparing}. These debates are relevant both for theories of human cognition (i.e., in terms of identifying which behaviors can be attributed to linguistic input) and for understanding the limits of current LMs (i.e., what they can and can't do). 

Here, we focus on \textit{false belief reasoning}: the human capacity to reason about the belief states of others, even when those beliefs are different from one's own---a capacity sometimes grouped under the broader umbrella of ``Theory of Mind''. This capacity has been widely studied in humans \citep{apperly2012theory}, non-human animals \citep{call2008does}, and more recently, language models \citep{kosinski2023theory, shapira-etal-2024-clever}. Recent work has generated debate about whether psychological tasks appropriately measure mental state reasoning in LMs \citep{trott2023large, ivanova2025evaluate}, with demonstrations that LM behavior is not robust to small stimulus perturbations \citep{ullman2023large, shapira-etal-2024-clever}. One less discussed---though equally crucial---limitation of much published work on this topic concerns the \textit{LM sample} typically used: empirical investigations often rely on a \textbf{relatively \textit{small sample} of mostly \textit{closed-source} LMs}---for instance, \citet{trott2023large} evaluate a single closed-source model (\textit{text-davinci-002}), which has since been deprecated.

This limitation presents several major inferential challenges. First, as others have argued \citep{liesenfeld2024rethinking, hussain2024tutorial, conde2025adding}, the use of closed-source models makes it virtually impossible to detect \textit{data contamination}, calling the validity of claims based on those evaluations into question. A related challenge is that most providers of closed-source models do not make public even relatively coarse-grained properties such as the size of a model, the number of tokens on which it was trained, or whether (and how) it was instruction-tuned---making it difficult to draw inferences about \textit{which properties} of a learner affect its downstream behavior. Moreover, closed-source models are vulnerable to deprecation, raising challenges for scientists attempting to \textit{reproduce} previously published work. Finally, beyond the limitations of closed-source models specifically, the small sample size used in most published work makes it difficult to determine the \textit{generalizability} of empirical findings obtained on a handful of model instances \citep{trott2025toward}. 

These concerns are not ``merely'' methodological: they represent fundamental \textit{epistemological challenges} for a systematic study of LM capacities. Building generalizable scientific theories of LMs requires studying representative samples of model instances about which relevant properties (e.g., model size) can be verified. Further, the use of LMs as ``model organisms'' to test theories about human cognition depends on accurate knowledge about those LMs. For example, knowing the number of words on which an LM was trained is extremely relevant for testing theories about the role of \textit{language exposure} in the emergence of particular capacities like false belief reasoning \citep{brown_1996_WhyTalkMental, dove2020more}.

In the current work, we address this limitation by reproducing and extending a previously published analysis of human and LM behavior on the false belief task  \citep{trott2023large}, using a larger sample of open-weight (and some open-source) LMs from five model families. We first survey related work on the \textit{false belief task} (Section \ref{sec:related}). We then address several research questions. First, we reproduce previously published analyses \citep{trott2023large} with a larger sample of LMs, asking which open-weight LMs (if any) are sensitive to a character's knowledge state (Section \ref{results:sensitivity}), and whether any LMs fully ``explain away'' the equivalent human effect (Section \ref{results:baselines}). Second, we exploit the larger sample to ask which properties of an LM predict its sensitivity to knowledge states (Section \ref{results:factors}), as well as its psychometric predictive power (Section \ref{results:ppp}). Finally, we use the sample of LMs to generate (and test) a \textit{novel hypothesis} about human behavior on the false belief task involving the role of \textit{non-factive verbs} (e.g., ``think'') in belief attribution (Section \ref{subsec:factive_verbs}). These questions (and empirical findings) are summarized in Table \ref{tab:rq_summary} below. Finally, we conclude by discussing the implications of these results for the study of mental state reasoning in LMs and humans (Section \ref{sec:discussion}), as well as key limitations and directions for future work (Section \ref{sec:limitations}). Note that all materials, data, and code necessary to reproduce these results will be published on GitHub when the anonymity period is over.

\begin{table}[h]
\centering
\small
\renewcommand{\arraystretch}{1.5}
\begin{tabular}{p{0.6\linewidth}c}
\hline
\textbf{Research Question} & \textbf{Finding} \\
\hline
RQ1: Are any open-weight LMs sensitive to manipulations of knowledge state? & \checkmark \\
RQ2: Do any open-weight LMs ``explain away'' the human effect? & \ding{55} \\
RQ3: Which LM properties correlate with sensitivity to knowledge state? & Parameter count \\
RQ4: Which LM properties correlate with psychometric predictive power? & Parameter count \\
RQ5: Are LMs and humans more likely to attribute false beliefs in the presence of non-factive verbs like ``thinks''? & \checkmark \\
\hline
\end{tabular}
\caption{Summary of research questions and findings.}
\label{tab:rq_summary}
\end{table}

\section{Related Work}\label{sec:related}

\subsection{False Belief Reasoning in Human Populations}

The capacity to reason about mental states has been extensively studied in humans, often using a version of the \textit{false belief (FB) task} \citep{wimmer1983beliefs, bradford2020neural}. In the canonical FB task, a participant witnesses the transfer of an object from one location (the ``start'') to a new location (the ``end''), with or without the knowledge of another character. The participant must then report (either directly or indirectly) where the character likely thinks the object is. While the validity of the FB task has been criticized on various grounds \citep{bloom_2000_TwoReasonsAbandon, gernsbacher2019empirical, hayward2017reliability}, it is widely used to assess mental state reasoning in humans \citep{bradford2020neural, fairchild2021role, schneider2014implicit}, and has even been adapted to study mental state reasoning in non-human animals \citep{premack_1978_DoesChimpanzeeHave, call2008does, krupenye_2016_GreatApesAnticipate, halina2015there} (see also Section \ref{sec:discussion}). Most relevantly, there is empirical support for the theory that \textit{language exposure} can augment mental state reasoning capacity \citep{devilliers_2014_RoleLanguageTheory, de2007interface, brown_1996_WhyTalkMental, gola2012mental, hale_2003_InfluenceLanguageTheory}. While not all of this evidence is \textit{causal}, it does point to a correlation between types of linguistic knowledge and the capacity to reason about belief states.

\subsection{False Belief Reasoning in Language Models}\label{subsec:fb_review_lm}

A growing body of research has investigated the ability of language models (LMs) to produce responses that are \textit{sensitive} to manipulations of a character's belief states \citep{sap-etal-2022-neural}, often using a version of the FB task \citep{kosinski2023theory, trott2023large, shapira-etal-2024-clever}. In some cases, this has motivated the creation of Theory of Mind benchmarks \citep{wu-etal-2023-hi, xu-etal-2024-opentom, jones2024comparing, gandhi2023understanding, gu2024simpletom}. Although results vary across tasks and models, a common finding is that LMs exhibit \textit{some} sensitivity to canonical belief-state manipulations \citep{kosinski2023theory}, but this sensitivity falls short of human performance \citep{trott2023large} and breaks down under minor stimulus perturbations \citep{ullman2023large, shapira-etal-2024-clever}. As discussed in Section \ref{sec:limitations}, there is considerable debate about the interpretation of these findings \citep{hu2025re}, particularly with respect to the construct validity of the FB task for LMs \citep{trott2023large, ivanova2025evaluate}. 

An equally important limitation, however, is that it often relies on a small sample of closed-source LMs to draw claims about mental state reasoning capacity in LMs more generally. This is particularly problematic when the goal of research is identifying the properties of individual learners that predict FB task performance---or evaluating the \textit{sufficiency} of distributional language statistics in accounting for human behavior on the FB task. 

\subsection{Motivation}\label{sec:motivation}

To address this limitation, we selected \citet{trott2023large} as a target for replication and extension for several reasons: first, the original study collected novel human data as a comparison for LM behavior, and explicitly investigated the link between distributional statistics and human FB task performance; second, the study used entirely novel stimuli, reducing the possibility of data contamination; third, the study used only a single model instance (\textit{text-davinci-002}), which has since been deprecated; and fourth, the study analyzed (and made public) individual human data, allowing us to predict trial-level results rather than solely comparing to an average.

Notably, the current work is also (to our knowledge) a novel approach to testing the distributional hypothesis as an explanation of human behavior on a psycholinguistic task: by evaluating a \textit{range} of LMs, we can directly quantify the likelihood of obtaining a given human effect size relative to the distribution of LM effect sizes (see Figure \ref{fig:prop_start2}).

\section{Method}\label{sec:method}

\subsection{Materials}

We used the same written passages developed by \citet{trott2023large}, which followed the general structure of the canonical False Belief (FB) Task (see Figure \ref{fig:prompt.example} for an example). Specifically, in each scenario, a character places an object in the Start location (e.g., a box), and a second character moves that object to the End location (e.g., a basket). The key manipulation of Knowledge State corresponds to whether the first character is present (True Belief) or not present (False Belief) when the object is moved. Each passage ends with a probe statement assessing the character's belief about the object's location, either Explicitly (e.g., ``John believes the book is in the...'') or Implicitly (e.g., ``John looks for the book in the ...''). The authors also manipulated which location (e.g., the box vs. the basket) first in the passage, and which was mentioned most recently. Each location was always mentioned twice in the passage. There were 12 different scenarios, with 16 versions each (Knowledge State x Knowledge Cue x First Mention x Recent Mention), for a total of 192 passages. 

\subsection{Models}

We aimed to sample a diversity of model sizes, training regimes, and providers. We selected a total of 41 models from 5 model \textit{families} \citep{gemma_team_gemma_2024,grattafiori_llama_2024,olmo_team_2_2025,biderman_pythia_2023,qwen_qwen25_2025}, where ``family'' refers to a collection of model instances released by the same model provider; these model instances typically share a number of features (such as pre-training methodology), but also vary in key respects (such as number of parameters, or whether a model is instruction-tuned).\footnote{See Appendix \ref{sec:appendix_api} for an analysis of additional models and model families accessed through the HuggingFace Inference Endpoints API.} 

\begin{table}[h]
\centering
\small
\begin{tabular}{lll}
\hline
\textbf{Family} & \textbf{Huggingface ID} & \textbf{Versions} \\
\hline
Gemma 2 & google/gemma-2-2b* & Base, Instruct \\
Llama 3 & meta-llama/Meta-Llama-3-8B* & Base, Instruct \\
OLMo-2 & allenai/OLMo-2-*-1124 & Base, Instruct \\
Pythia & EleutherAI/pythia-* & Base only \\
Qwen 2.5 & Qwen/Qwen2.5-* & Base, Instruct \\
\hline
\end{tabular}
\caption{Model families tested in our experiments. See Appendix \ref{sec:appendix_api} for an additional results with models run through the HuggingFace Inference Endpoints API.}
\label{tab:models}
\end{table}

\subsection{Procedure}

Model instances were run locally on an NVIDIA DGX-H200 using the HuggingFace \textit{transformers} package \citep{wolf-etal-2020-transformers}; see Appendix \ref{sec:appendix_api} for a supplementary analysis using larger models accessed through the HuggingFace Inference Endpoints API.

Each model instance was assessed in the following way. A stimulus was first \textit{tokenized} using that model's tokenizer. Following \citet{trott2023large}, we then presented the tokens to the model and extracted the log probability corresponding to the Start location token (e.g., ``box'') and the End location token (e.g., ``basket'').\footnote{Multi-token words were addressed by summing the log probabilities across all constituent tokens.} We then calculated the Log Odds by taking the difference between these log probabilities, such that a positive Log Odds indicated a higher probability was assigned to the Start location token. This was repeated for all 192 passages, for each model instance.

For instruction-tuned models, we carried out an additional analysis in which the models were prompted to provide an answer to a question (as opposed to simply completing a sentence), which was intended to mirror the instruction-tuning process (see Appendix \ref{sec:appendix_prompting}).

\subsection{Human Data}

We used publicly available data from \citet{trott2023large}. After applying exclusion criteria, a total of 613 participants completed the original single-trial experiment, which involved reading the target passage and completing the final sentence, e.g., ``John thinks the book is in the ...'. These were coded in terms of whether the participant responded with the Start or End location.

\section{Results}\label{sec:results}

All analyses were conducted in R \citep{r_cite}. Mixed effects models were constructed using the \textit{lme4} package \citep{lme4_cite}, and visualizations were produced using \textit{ggplot2} \citep{ggplot}. 

\subsection{RQ 1: Are any open-weight LMs sensitive to Knowledge State?}\label{results:sensitivity}

We first asked whether any of the open-weight models we tested were \textit{sensitive} to implied belief states. Specifically, as in past work \citep{trott2023large}, we asked whether the distribution of Log Odds produced by a given model instance varied systematically according to the manipulation of Knowledge State. For each model instance, we constructed a linear mixed effects model with Log Odds as a dependent variable; fixed effects of Knowledge State, Knowledge Cue, First Mention, and Recent Mention; and by-item random slopes for the effect of Knowledge State. We then compared this model to a model omitting only the effect of Knowledge State using a likelihood ratio test (LRT). 

\begin{figure}
\centering
\includegraphics[width=0.9\linewidth]{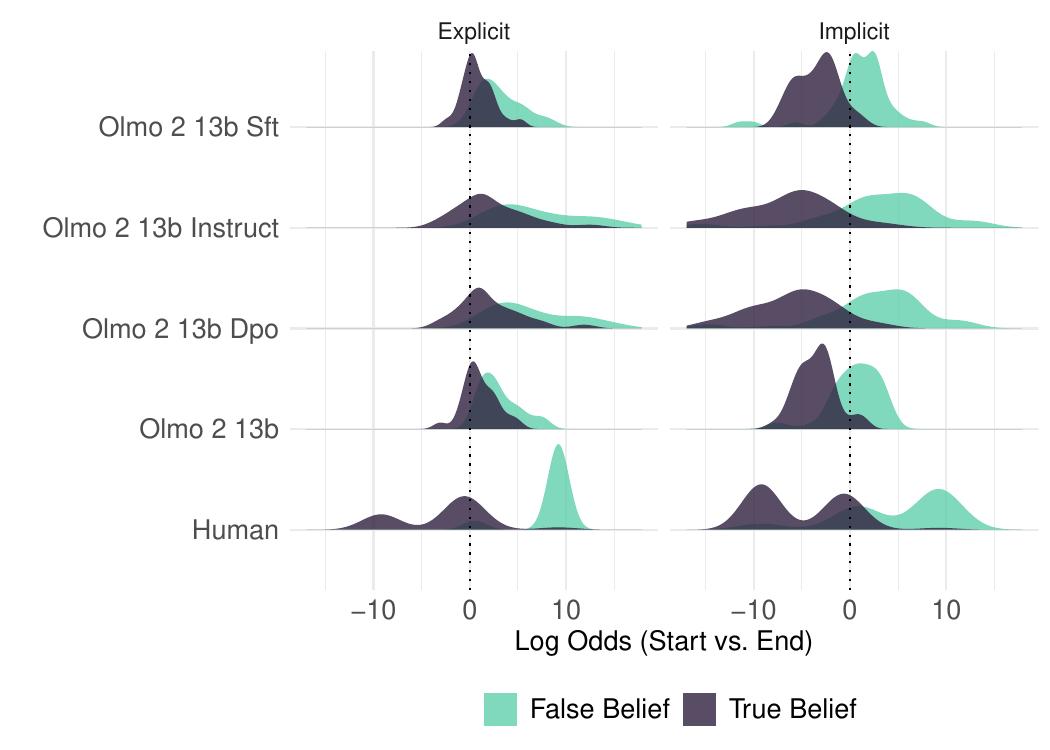}
\caption{Distribution of Log Odds by condition (True Belief vs. False Belief) for the best-performing models, along with the by-item human baseline. Human Log Odds were calculated by comparing the proportion of Start vs. End responses for each item across participants.}
\label{fig:sensitivity}
\end{figure}

After applying Holm-Bonferroni corrections, 14 of the 41 models tested (approximately $34\%$) exhibited sensitivity to Knowledge State (see Figure \ref{fig:sensitivity} for a subset of these models). In each case, the effect was in the right direction: a more negative Log Odds was observed for the True Belief condition, corresponding to a (relatively) higher probability assigned to the End (correct) location. Likelihood-ratios for this subset of models ranged from $11.9$ to $21.4$. We also conducted a follow-up analysis in which all LMs were analyzed together in a single mixed effects model, with by-LM random slopes for the effect of Knowledge State. Again, this full model explained significantly more variance than a model omitting only the effect of Knowledge State $[\chi^2(1) = 14.78, p < .001]$, and more negative Log Odds were again observed in the True Belief condition $[\beta = -1.5, SE = 0.37, p < .001]$.

Viewed from the lens of \textit{accuracy}---positive Log Odds for False Belief passages, and negative Log Odds for True Belief passages---the top-performing LMs achieved $74.5\%$ accuracy, roughly analogous to the performance of GPT-3 in past work. The best performance was achieved by instruction-tuned variants of the OLMo 2 13B model, as well as the base model. Average accuracy across the entire sample of LMs was $56.4\%$.

\subsection{RQ 2: Do any open-weight LMs ``explain away'' the  effect on humans?}\label{results:baselines}

Past work found that LMs did not fully ``explain away'' the effect of Knowledge State on human responses. We replicated this ``distributional baselines'' analysis for each LM by constructing a generalized linear mixed effects model with Human Response as a dependent variable (Start vs. End); fixed effects of Knowledge State, Log Odds, Knowledge Cue, Recent Mention, and First Mention; and random intercepts for each item. We compared this full model to a model omitting only the effect of Knowledge State to determine whether the manipulation explained variance in behavior beyond Log Odds. For all LMs, the full model explained significantly more variance (likelihood ratios ranged from $123.6$ to $195.4$). That is, Log Odds were \textit{insufficient} to fully account for the effect of Knowledge State on human behavior. 

More straightforwardly, human accuracy (approximately $82.7\%$) surpassed even the top-performing LMs ($74.5\%$) (see Figure \ref{fig:accuracy_params}). We also calculated an approximate effect size by comparing the proportion of Start responses across False Belief and True Belief conditions---a difference that should be positive, since characters are more likely to look in the Start (incorrect) location in the False Belief condition. As depicted in Figure \ref{fig:prop_start2}, this effect was considerably larger for humans than for any LM tested. Thus, the distributional statistics of language---as operationalized by this sample of LMs---account for \textit{some sensitivity} to manipulations of Knowledge State, but do not currently account for human-level sensitivity.

\subsection{RQ 3: Which factors predict LM performance?}\label{results:factors}

A key advantage of testing a large sample of LMs, as opposed to a single instance, is that it allows researchers to investigate which properties of individual model instances predict variance in performance. We focused on three factors found to be relevant to performance on various tasks in past work  \citep{kaplan2020scaling, ouyang2022training}: the number of parameters in a model, the number of tokens a model was trained on, and whether a model was subject to instruction-tuning. 

\begin{figure}
    \centering
    \includegraphics[width=0.9\linewidth]{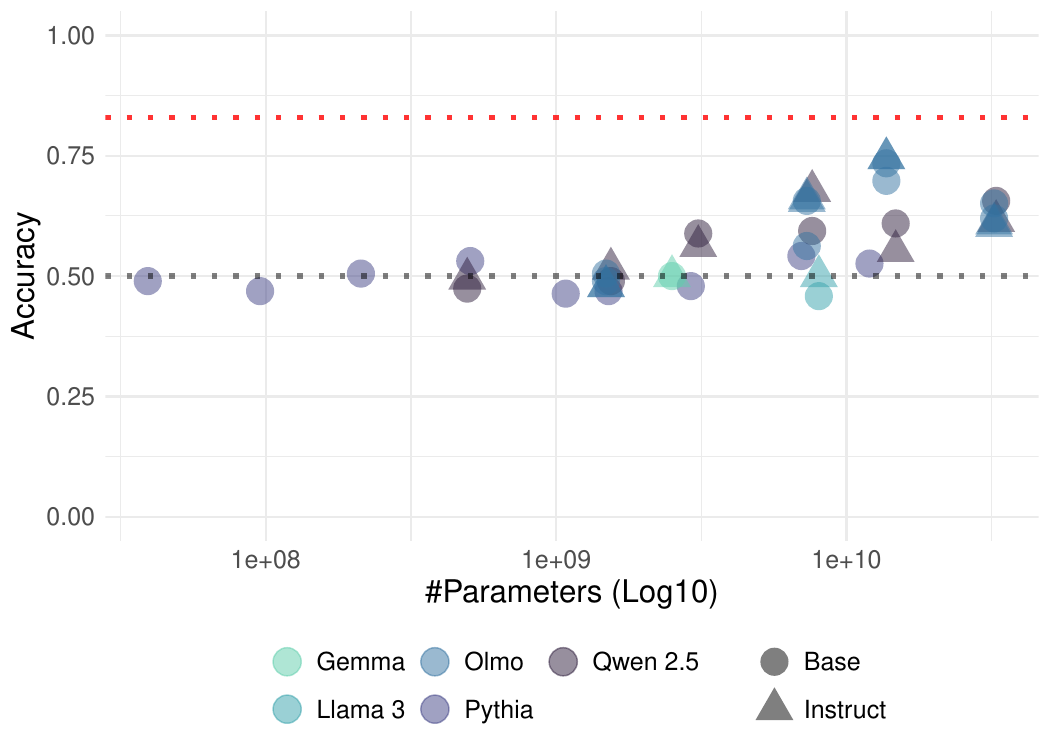}
    \caption{Accuracy on the false belief task by number of parameters. Overall, larger models performed better, though no model attained or surpassed the human baseline. Dashed red line corresponds to average human performance.}
    \label{fig:accuracy_params}
\end{figure}

These factors were entered as predictors in a linear regression model with mean accuracy as a dependent variable (parameter count and token count were both log-transformed). Altogether, this model achieved an $R^2 = 0.44$, though only Log Number of Parameters had a significant (positive) coefficient $[\beta = 0.08, SE = 0.02, p < .001]$: that is, each order of magnitude increase in parameter count was associated with an expected increase of about $8\%$ in accuracy (see also Figure \ref{fig:accuracy_params}). Of course, this relationship was not perfectly monotonic: the best-performing model (OLMo 2 13B) was about half the size of the largest models tested (e.g., OLMo 2 32B). Note that we also carried out additional analyses of instruction-tuned variants to investigate the role of specific prompt formatting; these results are described in Appendix \ref{sec:appendix_prompting}.

\subsection{RQ 4: Which factors contribute to psychometric predictive power?}\label{results:ppp}

\begin{figure*}
    \centering
    \begin{subfigure}[t]{0.48\textwidth}
        \centering
        \includegraphics[width=\textwidth]{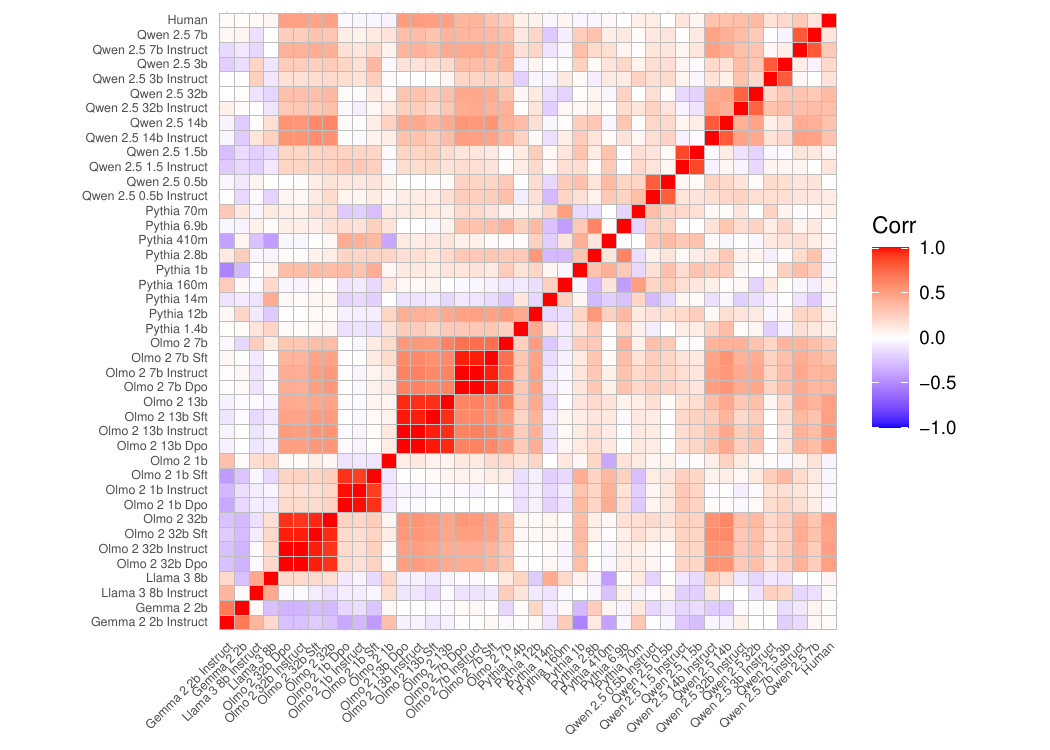}
        \caption{Correlation matrix between model outputs and human responses. Human responses were aggregated to create a Log Odds measure for each item.}
        \label{fig:corr_matrix}
    \end{subfigure}
    \hfill
    \begin{subfigure}[t]{0.48\textwidth}
        \centering
        \includegraphics[width=0.9\linewidth]{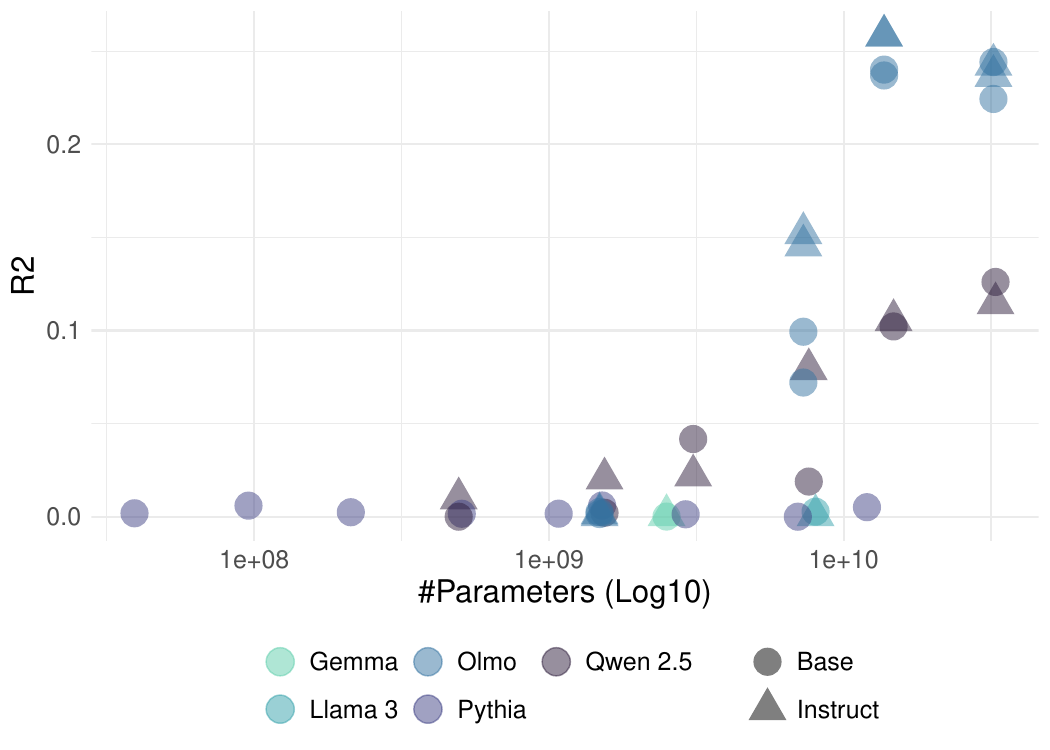}
        \caption{Psychometric predictive power (PPP) for each LM instance, as determined by the proportion of variance explained ($R^2$) in by-item human Log Odds by that LM's Log Odds. Larger models achieved higher PPP.}
        \label{fig:r2_params}
    \end{subfigure}
    \caption{Although no LM fully ``explained away'' the effect of Knowledge State in humans, some LMs did produce behavior that was more correlated with human behavior overall.}
    \label{fig:model_analysis}
\end{figure*}

Another metric of performance is \textit{psychometric predictive power} (PPP), i.e., the extent to which a given LM's behavior \textit{predicts} human behavior \citep{wilcox2023language}. Although no LM ``explained away'' the effect of Knowledge State on human responses (see RQ 2), LMs did vary considerably in their performance, leading us to ask whether some LMs were more correlated with human behavior than others.

We calculated PPP in two ways. First, we asked how much variance in human behavior could be accounted for as a function of LM Log Odds---setting aside other potential explanatory variables such as Knowledge State. For a direct comparison, we aggregated across human responses for each item to compute a Log Odds measure (i.e., the log ratio of the proportion of participants choosing the Start location vs. the proportion of participants choosing the End location). We then calculated the Pearson's correlation between each LM's Log Odds and the human Log Odds, as well as each pair of LMs (see Figure \ref{fig:corr_matrix}). PPP for each LM was approximated by calculating $R^2$, i.e., the proportion of variance in by-item Log Odds explained by that LM's Log Odds. The best-performing LM (an instruction-tuned variant of OLMo 2 13B) achieved an $R^2 = 0.26$. To ask which factors contributed to variance in PPP, we fit a multiple regression model predicting PPP from Log Parameter count, Log Training Tokens, and whether an LM was instruction-tuned: only Log Parameter Count was a significant predictor $[\beta = 0.09, SE = 0.02, p < .001]$ (see also Figure \ref{fig:r2_params}). 

In a second approach ($PPP_2$), we asked which LMs produced behavior that \textit{explained additional variance} above and beyond Knowledge State. We fit a generalized linear mixed effects model predicting human response (Start vs. End location) with fixed effects of LM Log Odds, First Mention, Recent Mention, Knowledge State, and Knowledge Cue, as well as random intercepts for each item. We compared the Akaike Information Criterion ($AIC$) of this model to the $AIC$ of a model omitting only LM Log Odds. Using this approach, a lower (more negative) $\Delta AIC$ corresponded to higher PPP relative to the other predictors, including Knowledge State. $\Delta AIC$ values ranged from $2$ (indicating no improvement in model fit, along with a penalty for the additional Log Odds parameter) to approximately $-8$. Lower $\Delta AIC$ was generally achieved by the models with the highest accuracy, though the two were not perfectly correlated (e.g., OLMo 2 7B obtained the lowest $\Delta AIC$). PPP was significantly predicted by Log Parameter Count in the expected direction, i.e., larger models achieved a relatively larger reduction in AIC $[\beta = -1.57, SE = 0.68, p = 0.03]$.

Together with RQ2, these results suggest that larger models were \textit{better at the FB Task} (RQ2) and better at accounting for human behavior on the FB task, as well as human behavior \textit{independent} of the manipulation of Knowledge State.

\subsection{RQ 5: Can LM behavior be used to generate novel hypotheses?}\label{subsec:factive_verbs}

One proposed benefit of using LMs in psycholinguistic research is the possibility of generating and testing \textit{novel} hypotheses \citep{misra2024generating, lakretz2021mechanisms}. Past work \citep{trott2023large} observed an effect of Knowledge Cue on GPT-3 Log Odds: namely, Log Odds were systematically higher in the Explicit condition (``John thinks the book is in the...'') than the Implicit condition (``John looks for the book in the...''). In other words, GPT-3 displayed a bias towards the \textit{incorrect} (Start) location when a \textit{non-factive} verb (``thinks'') was used. 

Although \citet{trott2023large} did not expand on this finding, there is a large body of work exploring the different presuppositions and entailments derived from factive and non-factive verbs \citep{hooper1975assertive, kiparsky1970fact}. Indeed, some \citep{chemla2008epistemic} have suggested that non-factive verbs (e.g., ``think'' or ``believe'') carry an \textit{anti-presupposition}: that is, ``John \textit{thinks} A'' may weakly imply \textit{NOT A}. This would be broadly consistent with the finding described above (i.e., imputing an incorrect belief to an agent when a non-factive verb is used), even though the original study was not designed with this hypothesis in mind. Moreover, there is also some evidence that LMs are sensitive to epistemic cues, such as expressions of uncertainty \citep{zhou-etal-2023-navigating}.

We extended the original finding by asking whether any LMs in the current sample also exhibited sensitivity to Knowledge Cue, whether humans showed analogous sensitivity, and whether the human effect was larger or smaller than the LM effect. For more direct comparison of the coefficients, we converted each LM's Log Odds into a categorical measure reflecting whether the ratio was positive (Start-biased) or negative (End-biased). Then, for each LM, we constructed a generalized linear mixed effects model predicting Response (Start vs. End), with fixed effects of Knowledge State, Knowledge Cue, First Mention, and Recent Mention (and by-item slopes of Knowledge State). We extracted the coefficient for Knowledge Cue for each LM: the mean across all LMs tested was negative $\bar{\beta}_{\text{KC}} = -1.6$ (median = $-1.08$, $SE = 0.43$\footnote{Standard error values were calculated by taking the standard deviation of the coefficient estimates across the entire sample and dividing by the square root of the sample size.}), and values ranged from $-11.67$ to $1.5$. A multiple regression model with $\beta_{KC}$ as a dependent variable, and predictor terms for Log Parameter Count, Log Token Count, and Base/Instruct status, indicated that larger models showed more negative (stronger) effects on average $[\beta = -2.04, SE = 0.63, p = 0.003]$. 

We then fit an analogous model for the human data, and the resulting coefficient for Knowledge Cue was also negative $[\beta = -1.02, SE = 0.27, p < .001]$: that is, humans were more likely to respond with the Start (incorrect) location in the Explicit condition, i.e., when the verb ``thinks'' was used (and more likely to respond with the End location in the Implicit condition). Together, these results suggest that both LMs and humans were biased towards reporting \textit{incorrect} belief states when non-factive verbs like ``think'' were used, perhaps because of an implied \textit{anti-presupposition} \citep{chemla2008epistemic}. The human parameter estimate fell in approximately the middle percentile ($49\%$) of LM parameter estimates. 

As with the primary Knowledge State effect, we visualized these effect sizes by comparing the probability of a Start response across Explicit vs. Implicit conditions (see Figure \ref{fig:prop_start1}). Here, a positive score reflects a bias towards the Start location in the Explicit condition (i.e., when a non-factive verb is used). In contrast to the effect of Knowledge State (where the human effect was much bigger than the distribution of LM effects), the effect of Knowledge Cue in humans fell squarely in the middle of the distribution of LM effect sizes. That is, distributional statistics are in principle \textit{sufficient} to account for the average effect size of Knowledge Cue on human responses, but not for the average effect size of Knowledge State.

\section{General Discussion}\label{sec:discussion}

\begin{figure*}
    \centering
    \begin{subfigure}[t]{0.48\textwidth}
        \centering
        \includegraphics[width=0.9\linewidth]{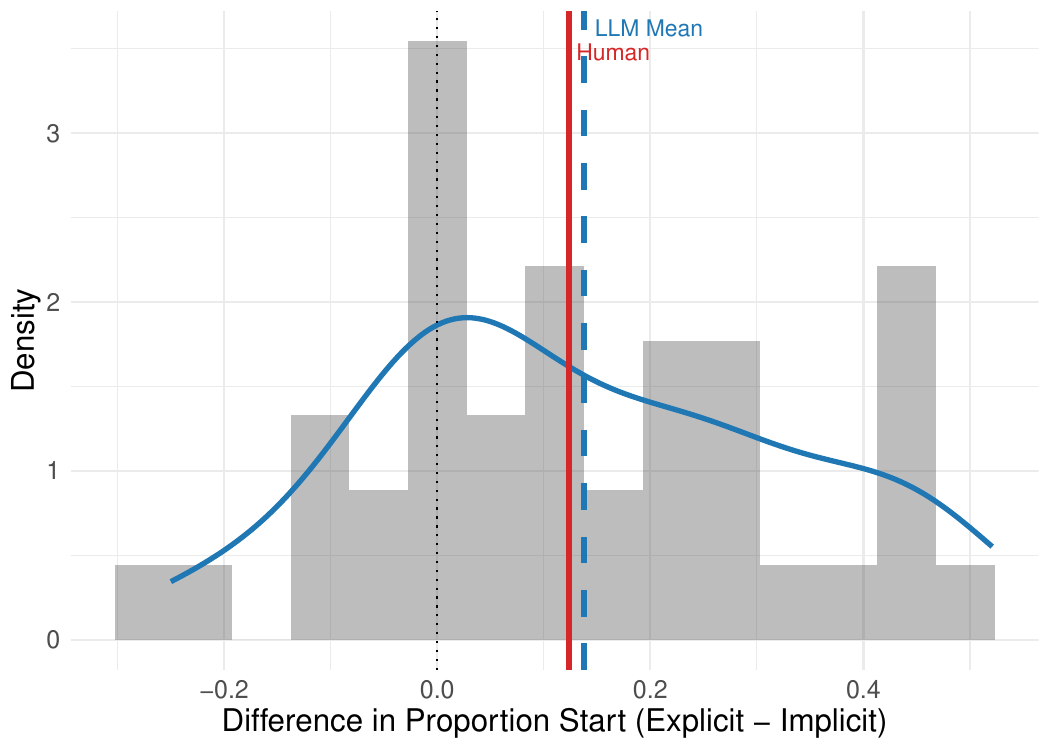}
        \caption{Distribution of differences in proportion of Start location responses across Explicit and Implicit conditions.}
        \label{fig:prop_start1}
    \end{subfigure}
    \hfill
    \begin{subfigure}[t]{0.48\textwidth}
        \centering
        \includegraphics[width=0.9\linewidth]{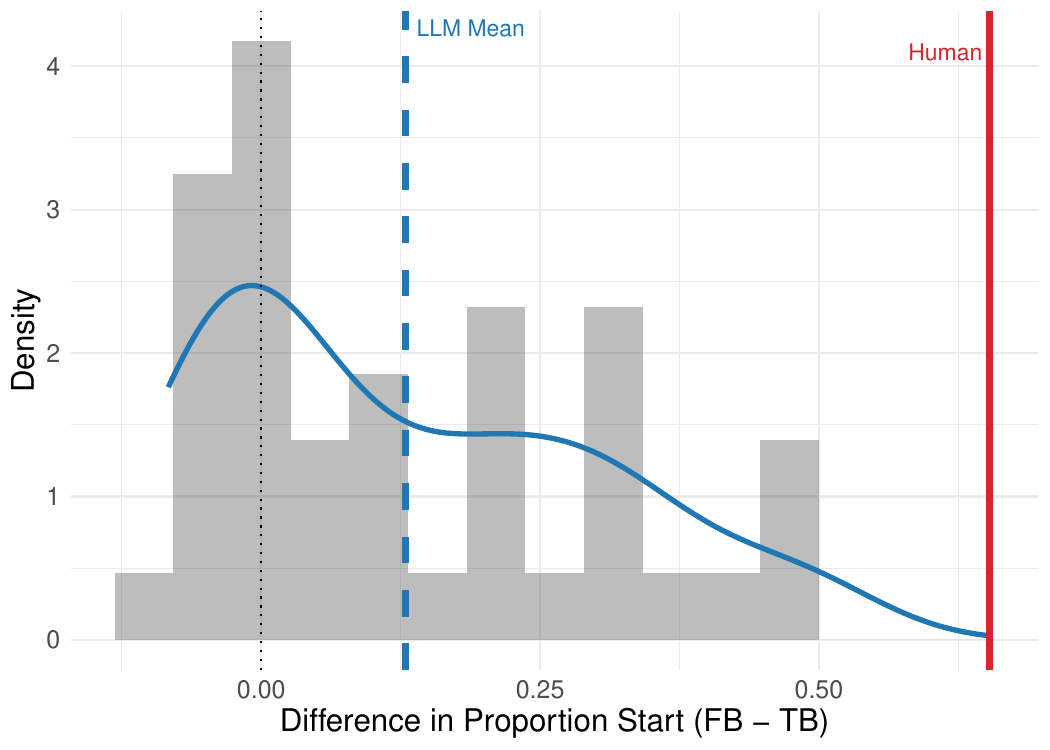}
        \caption{Distribution of differences in proportion of Start location responses across False Belief and True Belief conditions.}
        \label{fig:prop_start2}
    \end{subfigure}
    \caption{Both LMs and humans exhibited a bias towards the Start (incorrect) location when a non-factive verb (``thinks'') was used in the probe sentence. Both populations were also more likely to respond with the Start location in the False Belief condition, but this effect was considerably larger in humans.}
    \label{fig:prop_start_kc}
\end{figure*}

We tested a large sample of open-weight LMs (including a number of fully open-source LMs) on the \textit{false belief task} and used their behavior to predict individual human behavior on the same task. 
Our results point to several theoretical conclusions. First, LMs trained on the distributional statistics of language can develop \textit{sensitivity} to implied belief states, consistent with (though not entailing) the hypothesis that language exposure can partially account for mental state reasoning capacity in humans (Figure \ref{fig:sensitivity}). Moreover, larger LMs are more sensitive to mental states (Figure \ref{fig:accuracy_params}) and better predict human behavior on the same task (Figure \ref{fig:r2_params}). However, no LM fully ``explained away'' the human effect, suggesting that mental state reasoning in humans might well depend on other factors or processes beyond language statistics (Figure \ref{fig:prop_start2}). This is particularly striking given that the LMs tested are trained on orders of magnitude more language data than most humans encounter in a lifetime \citep{warstadt-etal-2023-findings}. That is, humans are more \textit{sample-efficient} in terms of the ratio of linguistic input to FB task performance (though not necessarily in terms of other input sources). The question of what accounts for this gap---grounding, social interaction, or even innate biological constraints---remains unknown, but could potentially be explored by systematically operationalizing each proposed mechanism in the context of LM training regimes or architectures.

The insufficiency of distributional statistics in accounting for manipulations of knowledge state also stands in sharp contrast to a novel finding reported in the current work: namely, LMs were (approximately) equally sensitive as humans to the manipulation of \textit{knowledge cue} (Figure \ref{fig:prop_start1}). One potential explanation for this asymmetry is that the ``anti-presupposition'' implicit in non-factive verbs like ``thinks'' is plausibly a lexical or conventionalized phenomenon: if ``X thinks P'' appears  in many cases where P is uncertain or even false, then the association between ``thinks'' and false beliefs could be learned through the distributional statistics of language---provided the LM has also learned to encode the likelihood of ``P'' being true. The FB task, on the other hand, may require constructing a \textit{situation model} \citep{zwaan1995construction} and representing the likely belief states of various entities in the discourse. This computation might be more challenging to learn through language statistics.

These results build directly on published work relying on (now-deprecated) closed-source LMs \citep{trott2023large}. The findings are also broadly consistent with a growing body of research suggesting that although LMs are surprisingly capable on mental state reasoning tasks \citep{kosinski2023theory, jones2024comparing}, their performance remains relatively brittle in the face of small perturbations \citep{shapira-etal-2024-clever} and does not match human behavior \citep{trott2023large}. As argued in Section \ref{sec:intro}, a key contribution of the current work is the use of a larger sample of open-weight (and open-data) LMs, which has several benefits. First, it makes the work more reproducible. Second, it allows us to investigate the \textit{generalizability} of claims about LM capabilities, as well as the factors contributing to variation across model instances, which is crucial for building a systematic science of LM behavior and capabilities; such findings can also help ``taxonomize'' LMs in terms of their behavioral similarity (see also Appendix \ref{sec:mds}). And third, it provides a \textit{distribution} of ``baseline'' effect sizes attributable to language statistics, as opposed to relying on a point estimate from a single LM---which in turn facilitates more robust statistical comparison to human effect sizes. For instance, we observed that a large subset of LMs showed human-level effects for the manipulation of Knowledge Cue but not Knowledge State (Figure \ref{fig:prop_start1} and Figure \ref{fig:prop_start2}). This latter approach may prove useful for future work evaluating the cognitive capacities of LMs or using LMs as ``model organisms'' to test (or generate) hypotheses about human cognition. 

\section{Limitations}\label{sec:limitations}

One limitation of the current work concerns the false belief task itself. As others have argued \citep{bloom_2000_TwoReasonsAbandon, gernsbacher2019empirical}, the task likely involves capacities other than mental state reasoning and may not accurately reflect ``Theory of Mind'' even in human samples. Inferences about mental state reasoning capacities are further complicated when applying the test to LMs, which may apply heuristics or ``short-cuts'' to solve the task based on patterns observed in the training data \citep{shapira-etal-2024-clever}---raising the question about whether the task (or any measure originally developed for humans subjects) has \textit{differential construct validity} for humans and LMs \citep{trott2023large, ivanova2025evaluate, saxon2024benchmarks}. Our aim here was not to resolve these important theoretical questions, but rather to reproduce and extend existing work using open-weight models and to ask which behaviors can emerge \textit{in principle} from exposure to language statistics. We refer interested readers to more detailed discussions of construct validity elsewhere \citep{ivanova2025evaluate, trott2023large, shapira-etal-2024-clever, hu2025re, bean2025measuring}, and note that addressing these challenges will likely require open-weight (and ideally open-data) models, for which relevant properties---such as training data composition---can be more accurately characterized. That said, future work would likely benefit from greater \textit{coverage} of possible scenarios (i.e., stimuli) requiring mental state reasoning to minimize the likelihood that good performance by LMs (or humans) is driven by ``shortcuts'' overfit to a particular stimulus structure. Future work could also adopt a mechanistic approach \citep{wu2025large} to ask whether consistent ``circuits'' are consistently recruited for a variety of scenarios involving mental state reasoning.

A second limitation is \textit{external validity}: we assessed a (relatively) large sample of open-weight LMs from multiple model families, but it is unclear whether this sample is \textit{representative} of distributional learners more generally, or what such a representative sample might look like \citep{constantinescu2025investigating}. The question of \textit{generalizability} (i.e., across architectures, training regimes, random seeds, and scale) has emerged as a key concern in recent work studying LM behavior and internal mechanisms  \citep{olah2020zoom, olahdreams, tigges2024llm, van2025polypythias, trott2025toward, michaelov2025language, zhao2025distributional}. While there has been some work attempting to ``taxonomize'' LMs \citep{yax2024phylolm}, it remains unclear which empirical findings we should expect to generalize from a given LM instance to another. We aimed to sample a large range of model \textit{sizes} and also considered both ``base'' and ``instruct'' versions of different models, but the space of \textit{possible} models is vast and largely unexplored; therefore, we limit our claims to the observed sample of LMs. One exciting direction would be to ask which properties allow researchers to predict whether two LMs will exhibit convergent or divergent behavior on the same task, perhaps allowing for a task-driven taxonomy of model instances (see also Appendix \ref{sec:mds}). 

Notably, the current sample was limited in that it excluded closed-source, state-of-the-art LMs. It is plausible that such LMs would attain better performance on the FB task---at the same time, the lack of transparency around their training process or composition (e.g., system prompts, use of external ``tools'', etc.) makes them less appropriate candidates for operationalizing theories about the sufficiency of distributional statistics specifically. Future work could benefit from exploring which kinds of modifications to the base LM architecture or training objective augment performance on the FB task; such research would also inform theories of the mechanisms and processes undergirding human mental state reasoning.

\section{Acknowledgements}\label{sec:acknowledgements}
The authors are grateful to Stephen J. Hanson for generously providing access to an NVIDIA DGX-H200. Pamela D. Rivi\`ere was supported by the UCSD Chancellor's Postdoctoral Fellowship. James Michaelov was supported by a grant from the Andrew W. Mellon foundation (\#2210-13947) during the writing of this paper.

\bibliography{custom}

\appendix

\section{Models run through Endpoints API}\label{sec:appendix_api}

Select larger models were accessed using the HuggingFace Inference Endpoints API, which allows users to access model outputs through a server hosted by HuggingFace with the same \textit{transformers} \citep{wolf-etal-2020-transformers} library used for the models that were run locally. Specifically, we ran Llama 3 70B (Base and Instruct), Qwen 2.5 72B (Instruct), Llama 3.1 70B (Base and Instruct), and Mixtral 8x7B (Base and Instruct). The same procedure was applied as for the models run locally (i.e., tokenizing the passage and Log Odds of completing the passage with the Start vs. End token). We then replicated the primary analyses above.

First, all models tested were \textit{sensitive} to manipulations of knowledge state, even after adjusting for multiple comparisons (likelihood ratios $> 6$ in all cases); accuracy ranged from $57.8\%$ (for Mixtral 8x7B Instruct) to $73.4\%$ (for Llama 3 70B Instruct). Second, no model fully ``explained away'' the effect of Knowledge State in humans: that is, a mixed model containing a fixed effect of Knowledge State consistently explained variance in human responses above and beyond Log Odds from each of the LMs tested (likelihood ratios ranged from $117.6$ to $174.5$). Finally, all models showed a significant negative effect of Knowledge Cue, i.e., passage completions were biased towards the End (correct) location in the Implicit condition---consistent with the effects reported in the primary manuscript.

\section{Assessing instruction-tuned models with question prompts}\label{sec:appendix_prompting}

Past work on instruction-tuned models has generally found a performance benefit of instruction-tuning \citep{ouyang2022training}, including on the False Belief task \citep{trott2023large}. One potential limitation of the approach adopted in the primary manuscript is that the same prompt strategy was used for ``base'' models and for instruction-tuned variants, even though instruction-tuning typically involves training models to follow explicit prompts and generate responses to queries, rather than computing next-token probabilities for sentence completions. Although this was an intentional feature of our experimental design (to control for the exact prompt used), this could unfairly bias the results \textit{against} instruction-tuned models. To address this limitation, we conducted a follow-up study on the subset of instruction-tuned models using a prompt format that mirrored the approach taken in their training process. 

\subsection{Procedure}

We selected the subset of $21$ instruction-tuned models used in the original analysis. Each passage was presented to each LM, embedded in an appropriate \textit{prompt template} based on the instruction-tuning procedure unique to each model, and rephrased into a question-answer format minimally different from its original phrasing. As in the original analysis, we then calculated the Log Odds of the Start vs. End token following the \textit{answer} portion of the prompt template. These different formats are shown in Figure \ref{fig:prompt.example}.

\begin{figure}[h]
\begin{tcolorbox}[listing box]
\begin{Prompt}
David and Marta go out to get some wine for the party. When they get home, David stores the wine in the garage and grabs a drink from the fridge. Then, David goes out to get some snacks. While David is gone, Marta decides the wine would be best cooled, so she moves the wine out of the garage and into the fridge. David returns home and wants to put out the wine. \textcolor{red}{David thinks the wine is in the} \textcolor{blue}{[MASK]}

\textbf{<|endoftext|><|user|>\textbackslash{}n}David and Marta go out to get some wine for the party. When they get home, David stores the wine in the garage and grabs a drink from the fridge. Then, David goes out to get some snacks. While David is gone, Marta decides the wine would be best cooled, so she moves the wine out of the garage and into the fridge. David returns home and wants to put out the wine. \textcolor{red}{David thinks the wine is in the}: \textbf{\textbackslash{}n<|assistant|>\textbackslash{}n}\textcolor{blue}{[MASK]}

\textbf{<|endoftext|><|user|>\textbackslash{}n}David and Marta go out to get some wine for the party. When they get home, David stores the wine in the garage and grabs a drink from the fridge. Then, David goes out to get some snacks. While David is gone, Marta decides the wine would be best cooled, so she moves the wine out of the garage and into the fridge. David returns home and wants to put out the wine. \underline{\texttt{Where does David think the wine is?}} \textbf{\textbackslash{}n<|assistant|>\textbackslash{}n}\textcolor{red}{David thinks the wine is in the} \textcolor{blue}{[MASK]}
\end{Prompt}
\end{tcolorbox}
\caption{An example stimulus, formatted for continuation mode (top), instruction mode (middle), and instruction mode in a question-answer format (bottom). The \textcolor{red}{text in red} remains the same, but changes position slightly in the question-answer format. The \textcolor{blue}{blue [MASK]} is where the probe of the Start and End token occurs. \textbf{Text in bold} are tokens introduced through the chat template for instruction-tuned models. In this example, the OLMo chat template is shown. These tokens will look different depending on the specific chat template unique to each model. The \underline{underlined text} is the question probe used only in the question-answer format. }\label{fig:prompt.example}
\end{figure}

\subsection{Results}

On average, the instruction-tuned LMs achieved slightly higher performance when prompted using the Question strategy ($62\%$) vs. the default Completion strategy ($59.4\%$). A linear mixed effects model predicting the mean accuracy of each LM with Prompt Strategy (Question vs. Completion) and a random intercept for each LM revealed a significant effect of Prompt Strategy, with higher accuracy predicted in the Question condition $[\beta = 0.03, SE = 0.01, p = .03]$. Notably, this effect size was relatively small (an expected difference of about $3\%$); moreover, the top-performing LMs did not \textit{improve} when prompted using the Question strategy. Indeed, the maximum accuracy achieved was actually in the Completion condition ($74.4\%$, as reported in the primary manuscript). Together, these results suggest that the theoretical conclusions reached in the primary manuscript (e.g., that no LMs attain human-level performance on the FB Task) are not particularly dependent on the prompting strategy used for these instruction-tuned models.

\begin{figure}
    \centering
    \includegraphics[width=0.9\linewidth]{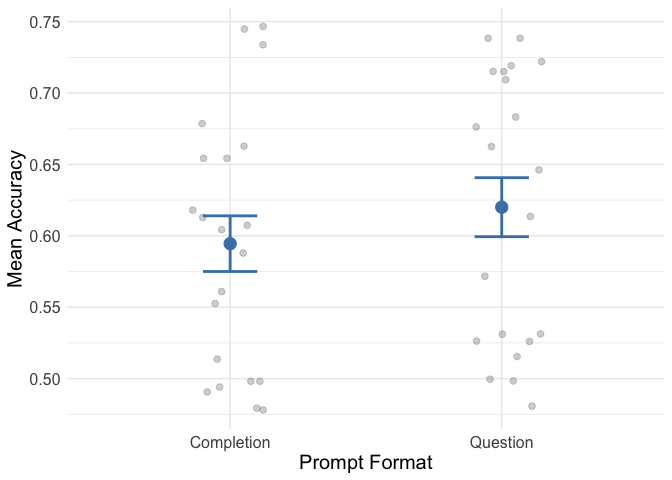}
    \caption{Mean accuracy was slightly higher when instruction-tuned models were using a question/answer (Question) format than when asked to complete the passage (Completion).}
    \label{fig:accuracy_by_prompting}
\end{figure}

\section{MDS analysis of models}\label{sec:mds}

\begin{figure}
    \centering
    \includegraphics[width=0.9\linewidth]{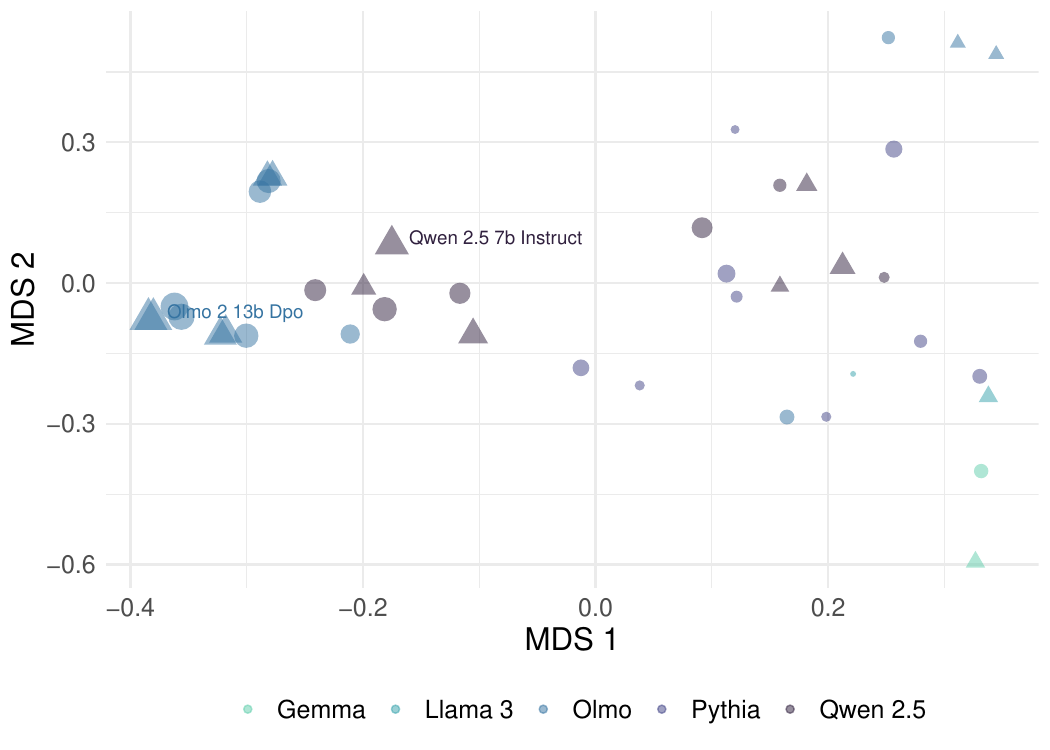}
    \caption{Results of multi-dimensional scaling (MDS) applied to the correlation matrix presented in the primary manuscript. Size of each dot indicates mean accuracy of that model instance; shapes represent Base/Instruct (circle = Base, triangle = Instruct).}
    \label{fig:mds}
\end{figure}

In the primary manuscript, we presented a correlation matrix illustration the Pearson's correlation between the behavior of each pair of models. Following other recent work in the study of larger LM samples \citep{huang2025measuring, trott2025toward}, we performed multi-dimensional scaling (MDS) on this correlation matrix to visualize how each individual clustered relative to other model instances. As depicted in Figure \ref{fig:mds}, model instances clustered strongly by \textit{model family}. Additionally, the first MDS component appeared to track \textit{accuracy}, with more accurate model instances receiving more negative scores. Together, this suggests that cross-model correlations can be explained both by overall performance (more accurate models are more similar to each other) and by aspects of the architecture and training procedure (instances from the same family cluster together).

\end{document}